\documentclass[conference]{IEEEtran}
\IEEEoverridecommandlockouts
\usepackage{cite}
\usepackage{amsmath,amssymb,amsfonts}
\usepackage{algorithmic}
\usepackage{graphicx}
\usepackage{textcomp}
\usepackage{xcolor}

\usepackage{times}
\usepackage{soul}
\usepackage{url}
\usepackage[hidelinks]{hyperref}
\usepackage[utf8]{inputenc}
\usepackage{graphicx}
\usepackage{amsmath,amsthm,amssymb}
\usepackage{algorithm}
\usepackage{algorithmic}
\usepackage{booktabs}
\usepackage{multirow}
\newcommand{\vct}[1]{\ensuremath{\mathbf{#1}}}

\newcommand{\set}[1]{\ensuremath{\mathcal{#1}}}
\newcommand{\con}[1]{\ensuremath{\mathsf{#1}}}

\newcommand{\argmin}{\operatornamewithlimits{\arg\,\min}}

\newcommand{\myparagraph}[1]{\smallskip \noindent \textbf{#1}}

\def\BibTeX{{\rm B\kern-.05em{\sc i\kern-.025em b}\kern-.08em
    T\kern-.1667em\lower.7ex\hbox{E}\kern-.125emX}}
\begin{document}

\title{The Hammer and the Nut: Is Bilevel Optimization Really Needed to Poison Linear Classifiers?}

\makeatletter
\newcommand{\linebreakand}{%
  \end{@IEEEauthorhalign}
  \hfill\mbox{}\par
  \mbox{}\hfill\begin{@IEEEauthorhalign}
}

\author{
\IEEEauthorblockN{Antonio Emanuele Cinà}
\IEEEauthorblockA{\textit{Ca' Foscari University of Venice} \\
\textit{DAIS}\\
Venice, Italy \\
antonioemanuele.cina@unive.it}
\and 
\IEEEauthorblockN{Sebastiano Vascon}
\IEEEauthorblockA{\textit{Ca' Foscari University of Venice} \\
\textit{DAIS}\\
Venice, Italy \\
sebastiano.vascon@unive.it}
\and
\IEEEauthorblockN{Ambra Demontis}
\IEEEauthorblockA{\textit{University of Cagliari} \\
\textit{DIEE}\\
Cagliari, Italy \\
ambra.demontis@diee.unica.it}
\linebreakand

\IEEEauthorblockN{Battista Biggio}
\IEEEauthorblockA{\textit{University of Cagliari} \\
\textit{DIEE}\\
Cagliari, Italy \\
battista.biggio@diee.unica.it}

\and
\IEEEauthorblockN{Fabio Roli}
\IEEEauthorblockA{\textit{University of Cagliari} \\
\textit{DIEE}\\
Cagliari, Italy \\
roli@diee.unica.it}

\and
\IEEEauthorblockN{Marcello Pelillo }
\IEEEauthorblockA{\textit{Ca' Foscari University of Venice} \\
\textit{DAIS}\\
Venice, Italy \\
pelillo@unive.it}
}

\maketitle

\begin{abstract}
One of the most concerning threats for modern AI systems is data poisoning, where the attacker injects maliciously crafted training data to corrupt the system's behavior at test time.
Availability poisoning is a particularly worrisome subset of poisoning attacks where the attacker aims to cause a Denial-of-Service (DoS) attack. However, the state-of-the-art algorithms are computationally expensive because they try to solve a complex bi-level optimization problem (the ``hammer''). 
We observed that in particular conditions, namely, where the target model is linear (the ``nut''), the usage of computationally costly procedures can be avoided. 
We propose a counter-intuitive but efficient heuristic that allows contaminating the training set such that the target system's performance is highly compromised. We further suggest a re-parameterization trick to decrease the number of variables to be optimized. Finally, we demonstrate that, under the considered settings, our framework achieves comparable, or even better, performances in terms of the attacker's objective while being significantly more computationally efficient.   
\end{abstract}


\section{Introduction}
%

The increasing pervasiveness of machine learning algorithms in high-stake real applications poses an issue about their robustness in the presence of adversarial manipulations. In \cite{barreno_security_2010}, the authors highlight two potential influence attacks that can deceive learning systems. 

On the one hand, \textit{exploratory} attacks exploit the target system's weaknesses to obtain misclassifications. Among them \textit{evasion attacks} have been getting a lot of attention in recent years \cite{biggio_evasion_2013, su_one_2019, carlini_towards_2017, goodfellow_explaining_2015, szegedy_properties}. In such attacks, the adversary alters the test samples to have them misclassified by the model. For example, an attacker may add a sticker \cite{song18-cvpr} to a stop sign to have it misclassified as another road sign, potentially causing a collision. Or slightly alter a malware to have it misclassified by anti-virus as a legitimate application \cite{demetrio20-arxiv-blackbox}. 

On the other hand, \textit{causative} attacks aim to influence the learning process by altering the training data, to meet the attacker's objective once the model is deployed. They are useful in applications where the attacker cannot directly manipulate test input because, for example, they are under surveillance \cite{shafahi_poison_2018} or when the attacker's goal is to cause the misclassification of all the received inputs generating a denial of service.

In a recent study, reported in \cite{industry_perspective}, poisoning attacks have caught more attention in companies, especially after that a Microsoft AI service was target by a malicious poisoning attack. In 2016 Tay, an artificially intelligent chat-bot developed by Microsoft's Technology and Research, started to twit inappropriate messages because of poisoning attack \cite{LearningfromTaysBlog}. Due to the large amount of data that machine learning systems consume, it is quite tricky to identify poisoning samples in a massive training set, leading Microsoft to switch-off the service.  Accordingly, to \cite{industry_perspective}, poisoning is nowadays considered by companies the attack against ML that would affect more their business. 

However, crafting a poisoning attack may be computationally expensive, according to the nature of the violation that the attacker desires. 
In \textit{integrity poisoning} the attacker aims to obtain the misclassification of a specific input or an exiguous amount of test samples with peculiar features. 
For instance, in \cite{chen_targeted_2017} a face recognition system is poisoned so that malicious users, wearing a particular pair of glasses, are classified as authorized ones. This attack can be performed with different efficient strategies \cite{shafahi_poison_2018, geiping_witches_2020, chen_targeted_2017}. Thanks to some heuristics, e.g., feature collision \cite{shafahi_poison_2018}, the attacker can also create integrity poisoning efficiently against systems trained on large-scale datasets.
Conversely, \textit{availability poisoning} attack aims to increase the test error causing a denial-of-service system. Nonetheless, its application has been limited in scope and success due to the computational requirements \cite{munoz-gonzalez_towards_2017, demontis_why_nodate, biggio_poisoning_2013, koh_understanding_nodate, yang_generative_2017}.

Mathematically speaking, availability (integrity) poisoning require solving a bi-level optimization problem where the outer problem consists of minimizing (maximizing) the accuracy on a validation set while reducing, in the inner problem, the accuracy on the poisoned training set \cite{geiping_witches_2020, munoz-gonzalez_towards_2017}.
Solving this problem with state-of-art strategies is computational costly since it requires re-training the model many times to generate the poisoning points.

In this work, we aim to bridge this gap, at least for simple classification problems. 
We found some interesting settings (the ``nuts'') where we don't need to solve a bi-level optimization problem (the ``hammer'') to create effective poisoning samples. In particular, we observed that linear classifiers can be corrupted efficiently with a carefully designed heuristic. 
Despite this classification problem's simplicity, efficient heuristics to perform availability poisoning does not exist yet, and the poisoning points are usually generated solving a complex and computationally demanding optimization problem. 
Therefore, we propose a new heuristic to efficiently perform availability poisoning. We further suggest a variable reduction technique to reduce the number of terms to be optimized during the learning process and consequently the computational effort.

In Section \ref{sec:threat_model} we first review the notion of attacker's knowledge and capability. Then we introduce the bi-level optimization problem, which formalizes the problem to find the optimal poisoning points. Subsequently, in Section \ref{sec:related_work} we review the related work about poisoning, and we discuss the actual limits of the state-of-art approaches. In Section \ref{sec:our_approach}, we present our approach to efficiently craft poisoning samples for availability violation. Moreover, we propose a variables reduction technique that can be used to further speed-up the optimization procedure. Finally, in Section \ref{sec:experiments} we test the proposed algorithm's effectiveness against support vector machines and logistic regression classifiers, both trained on MNIST and CIFAR-10. We make a performance and computational costs comparison against the existing poisoning algorithms designed against this family of models.

\section{Background} ~ \label{sec:threat_model}
In this section, we first explain the poisoning's threat model
\cite{biggio_wild_2018}, then we describe the bi-level problem that should be solved to generate the poisoning points needed to perform an availability poisoning attack.

\myparagraph{Notation.} Feature and label spaces are denoted in the following with $\set X \subseteq \mathbb R^{\con d}$ and $\set Y \in \mathbb R$, respectively, with $\con d$ being the dimensionality of the feature space. 
The attacker should collect a training and a validation/surrogate data sets that will be used to craft the attack. We denote them as $\set D_{tr}$ and $\set D_{val}$. We define $\set D_{\text{val}}^{y_t} = \{x | (x, y_t) \in \set D_{val}\} $ a subset of the validation set with samples of class $y_t$. Note that these sets include samples along with their labels.
We define with $L(\set D_{\rm val}, \vct \theta)$ the validation loss incurred by the classifier $f: \set X \rightarrow \set Y $, parametrized by $\vct \theta$, on the validation set $\set D_{\rm val}$. $\set L(\set D_{\rm tr}, \vct \theta)$ is used to represent the regularized loss optimized by the classifier during training.

\myparagraph{Attacker's Capability and Knowledge.}
Attacker's Capability and Knowledge are two fundamental characteristics that affect the strength of the attack that the attacker can create. 
The first defines how the attacker can affect the target system. If the attacker can manipulate both training and test data, the attack is said to be causative. Otherwise,  if the attacker can control only the test data, the attack is said to be exploratory. Besides, the attacker's capability also defines how data can be altered according to application-specific constraints. For example, to evade malware detection, the attacker should manipulate the code to fool the system without compromising its intrusive functionality.
The latter defines the information about the target system that the attacker can exploit to create the malicious threat. Characterizing the attacker's knowledge, we find the following components: the training data, the feature set, the learning algorithm, and its parameters.
Depending on the attackers' prior knowledge about these four components, the attack is defined as a white-, gray-, or black-box attack. In a white-box scenario, the attackers have complete knowledge about all the four components mentioned above. This scenario is the worst-case for the system defenders because, having more knowledge of their target, the attackers can craft stronger attacks.  
To fool a system in a black-box scenario, the attackers should firstly collect a surrogate dataset and train a surrogate model \cite{transfer_iclr17, carlini_towards_2017, transfer_cvpr19}. The attackers then craft the attack against the surrogate model and exploit it against the target system. 
The gray-box scenario represents a middle condition, where the attacker can leverage partial information about the four components. In \cite{demontis_why_nodate} the authors showed that availability poisoning attacks crafted in this setting can highly damage the target model if it has a similar complexity to the surrogate classifier used by the attacker.

\myparagraph{Poisoning Strategy.} \label{par:poison_strategy}
Using the notation mentioned above, we can formulate the optimal availability poisoning strategy in terms of the following bi-level optimization problem:
\begin{eqnarray}
    \label{eq:poisoning_problem_outer}
    \max_{\vct x_p} & &  L( \set D_{\rm val}, \theta^\star) \, , \\
    \label{eq:poisoning_problem_inner}
    {\rm s.t.} & &  \theta^\star \in  \argmin_{\theta} \set L (\set D_{\rm tr} \cup (\vct x_p, y_p), \vct \theta) \, , \\
    \label{eq:poisoning_problem_box}
    & & \vct x_{\rm lb} \preceq \vct x_p \preceq \vct x_{\rm ub} \, . 
\end{eqnarray}

The goal of this attack, developed in \cite{biggio_poisoning_2013, demontis_why_nodate, munoz-gonzalez_towards_2017}, is to maximize a loss function on a set of untainted (validation) samples by optimizing the poisoning sample with features $\vct x_p \in \mathbb R^{\con d}$ and label  $y_p$, as stated in the outer optimization problem Eq. (\ref{eq:poisoning_problem_outer}).
The poisoning sample is added to the training set $\set D_{\rm tr}$, used to learn the classifier solving the inner optimization problem Eq. (\ref{eq:poisoning_problem_inner}).
As one may note, the classifier $\vct \theta^ \star$ is learned on the poisoned training data and then used to compute the outer validation loss. This highlights an implicit dependency of the outer loss on the poisoning point $\vct x_p$ via the optimal parameters $\vct \theta^\star$ of the trained classifier. In other words, we can express the optimal parameters $\vct \theta^\star$ as a function of $\vct x_p$, i.e., $\vct \theta^\star(\vct x_p)$. This relationship tells us how the classifier parameters change when the poisoning point $\vct x_p$ is perturbed. Characterizing and being able to manipulate this behavior is the key idea behind poisoning attacks.
Eq. (\ref{eq:poisoning_problem_box}) imposes box constraints on the poisoning samples, where $\vct u \preceq \vct v$ means that each element of $\vct u$ has to be not greater than the corresponding element in $\vct v$.
As stated in \cite{geiping_witches_2020}, integrity poisoning can be similarly formulated as a bi-level optimization problem. In this way, the inner optimization problem, which represents the learning problem, remains the same. Instead, the outer optimization problem changes as integrity poisoning aims to reduce the classifier loss on the validation dataset.

\section{BetaPoisoning} ~ \label{sec:our_approach}
In the following, we describe our availability poisoning algorithm, named \textit{BetaPoisoning}. Notably, it does not need access to the training set. Neither it needs to re-train the target model during the optimization procedure.


Our goal is to exploit linear classifiers' limits when dealing with noisy-labeled samples \cite{DBLP:journals/datamine/Burges98}. In particular, we aim to poison the target distributions $y_t$ with sample $\vct x_p$ by maximizing the likelihood $P(\vct x_p | y_t)$, making the dataset no longer linearly separable.

\begin{figure}[t]
    \centering
    \includegraphics[width=0.45\textwidth]{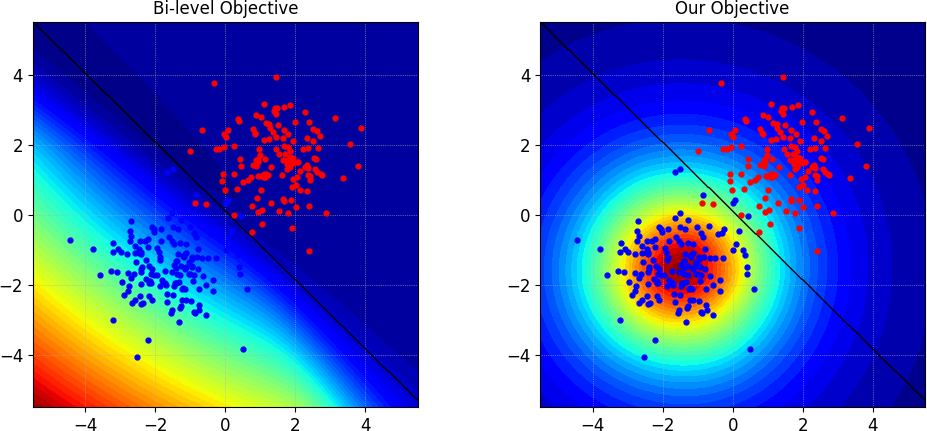}
  \caption{(Left) Attacker's objective function with bi-level formulation \cite{biggio_poisoning_2013}. (Right) Ours objective function, described in Eq. (\ref{beta_poisoning}). Red regions represent zones where the objective function is high. The black line is the decision boundary of a logistic regression classifier.}
  \label{fig:objective_function}
\end{figure}

We illustrate the idea behind the proposed approach with an example in Figure \ref{fig:objective_function}, which visualizes the difference between our objective function and the one optimized in Problem (\ref{eq:poisoning_problem_inner})-(\ref{eq:poisoning_problem_box}). To create an easily understandable example, we consider a linearly separable two-dimensional dataset in which each class follows a Gaussian distribution. 
Based on the bi-level problem illustrated in the previous section, the theoretical formulation suggests that the poisoning point should be located in the bottom-left region to obtain the highest validation error (left plot in Figure \ref{fig:objective_function}). The red area shows the optimal solution of the availability poisoning Problem (\ref{eq:poisoning_problem_inner})-(\ref{eq:poisoning_problem_box}). However, solving this problem is computationally expensive. 
Our heuristic approach, shown in the right plot of Figure \ref{fig:objective_function}, suggests locating the poisoning samples in the space region with the highest density of training samples. 
This is a counter-intuitive solution because the optimal region is quite different from the one obtained optimizing the bi-level problem.
The optimum suggested by our heuristic approach is distant from the optimal one. Nonetheless, optimizing the proposed objective function allows us to achieve good results under the considered settings.
This behavior may also occur for label flip attack \cite{taheri_defending_2020}, where the poison samples are randomly chosen from the validation dataset, and their labels get flipped. Notwithstanding, these points may be located in non-dense regions; hence even if we flip their labels, they may not significantly reduce the classifier's performance.

The formulation of the heuristic attack (BetaPoisoning) that we are proposing is the following:
\begin{eqnarray}
    \label{beta_poisoning}
    \arg\max_{\vct x_p} && P( \vct x_p | y_t)\\
    \label{eq:poisoning_problem_box_ours}
    {\rm s.t.}  && \vct x_{\rm lb} \preceq \vct x_p \preceq \vct x_{\rm ub} 
\end{eqnarray}
where Eq. (\ref{eq:poisoning_problem_box_ours}) defines box-constraints on the poisoning point's feature values. 
This heuristic attack considers only the data distribution; therefore, conversely to \cite{biggio_poisoning_2013, demontis_why_nodate} we don't need to know the model's parameters. 

To estimate the likelihood $P( \vct x_p | y_t)$ we use a Gaussian Kernel Density Estimator (KDE), where the bandwidth parameter $h$ is chosen equal to the average distance between all possible pairs of samples in $\set D_{val}^{y_t}$ \cite{battista_is_data_clustering}.

\begin{eqnarray}
    \label{eq:kde}
    P( \vct x_p | y_t) = \frac{1}{\|\set D_{val}^{y_t}\|} \sum_{\vct x \in \set D_{val}^{y_t}} \exp{\Big(-\frac{\|\vct x_p - \vct x\|^2}{h}\Big)}
\end{eqnarray}

To further decrease the computational complexity, we propose a re-parametrization trick that aims to reduce the number of variables to optimize during the learning process. We impose that our poisoning samples are obtained as a linear combination of other samples in the dataset. We define $\set S = \{ \vct x_1, \dots, \vct x_k \} $ a random subset of samples with label $y_t$ in $\set D_{val}$. We define with $k = |S|$ the number of samples, named prototypes, in $\set S$. Given a random set of samples $\set S$ and coefficients $\boldsymbol \beta \in \mathbb{R}^k$, the corresponding poisoning samples $\vct x_p$ is obtained with:

\begin{eqnarray}
    \vct x_p = \psi(\boldsymbol \beta, S) = \sum_{\vct x_i \in \set S} \beta_i \vct x_i 
\end{eqnarray}

We can thus reformulate our optimization problem in terms of $\boldsymbol \beta$ coefficients in the following way:

\begin{eqnarray}
    \label{trick_beta_poisoning}
    \arg\max_{\boldsymbol \beta} && p( \psi(\boldsymbol \beta, S) | y_t)\\
    {\rm s.t.} && \vct x_{\rm lb} \preceq \psi(\boldsymbol \beta, S) \preceq \vct x_{\rm ub} 
\end{eqnarray}
Once the optimal $\boldsymbol \beta$ coefficients are optimized we can easily reconstruct the resulting poisoning sample with $\psi(\boldsymbol \beta, S)$.
We observe that in our formulation for each poisoning point the $k$ prototypes are randomly sampled from the surrogate data. 

\myparagraph{Algorithm.}
A pseudo-code description of our attack can be found in Algorithm \ref{alg:beta_poisoning}. 
We get in input a surrogate dataset $\set D_\text{val}$, the target class $y_t$, the number of prototypes $k$, and the box constraints $(\vct x_{lb}, \vct x_{ub})$. We note that the attacker may choose $y_t$ and $y_p$ according to some strategy or prior knowledge about the application context. In our setting we assume that the attacker chose $y_t$ and $y_p$ randomly during the rest of this work. In line \ref{algorithm:init_proto} we randomly sample a subset of prototypes from $\set D_\text{val}$ and we initialize the corresponding $\boldsymbol \beta$ in line \ref{algorithm:init_beta}. Initialization for the $\boldsymbol \beta$ coefficients is done by randomly sampling values in $[0, 1]$. In line \ref{algorithm:build_poison} we construct the poisoning point and we clip it to preserve box constraints. In line \ref{algorithm:KDE} we estimate the likelihood $P(\vct x_p | y_t)$ with the Kernel Density Estimator in Eq. (\ref{eq:kde}). We then update the $\boldsymbol \beta$ coefficients through a gradient ascend step, with a learning rate $\alpha=0.01$. The process described from line \ref{algorithm:build_poison} to \ref{algorithm:gradient} is repeated until a certain stop condition is reached, i.e., if the attacker's objective function $P(\vct x_p | y_t)$, in two consecutive iterations, does not change more than a pre-defined threshold. 
In our experiments, we set this threshold to $1e-05$.

\begin{algorithm}[t]
 \caption{BetaPoisoning}\label{alg:beta_poisoning}
 \begin{algorithmic}[1]
   \STATE {\bfseries Input: $\set D_{val} \in \mathbb{R}^{n\times \con d}, y_t, k, \vct x_{lb}, \vct x_{ub}$}
   \STATE \textbf{Output:} Poison sample $\vct x_p \in \mathbb{R}^d $
   \STATE
   \STATE $\set S$ = \texttt{sample\_random\_prototypes}($\set D_{val},  y_t$)  \label{algorithm:init_proto}
   \STATE $\boldsymbol \beta$ = \texttt{init\_beta}(k) \label{algorithm:init_beta}

   \REPEAT 
        \STATE $\vct x_p$ = \texttt{clip}$(\psi(\boldsymbol \beta, S), ~\vct x_{lb}, ~\vct x_{ub})$ \label{algorithm:build_poison}
        \STATE $p = P(\vct x_p | y_t)$ \label{algorithm:KDE}
        \STATE $\boldsymbol \beta = \boldsymbol \beta + \alpha \nabla_{\boldsymbol \beta} p$ \label{algorithm:gradient}
   \UNTIL{{stop condition is reached}}
   \STATE $\vct x_p $= \texttt{clip}$(\psi(\boldsymbol \beta, S), ~\vct x_{lb}, ~\vct x_{ub})$
   \STATE {\bfseries return:} $\vct x_p$
\end{algorithmic}
\end{algorithm}

\section{Experimental Analysis} ~ \label{sec:experiments}
This section evaluates the effectiveness of our poisoning attack on two publicly available datasets, MNIST and CIFAR-10. 
We train and test the robustness of a linear support vector machine (SVM) and a logistic regression classifier (LC) under different regularization levels for both datasets. We then propose a computational comparison between our approach, the random label flips attack \cite{bengio_label_flip}, and the bi-level algorithms proposed in \cite{biggio_poisoning_2013} (for SVM) and \cite{demontis_why_nodate} (for LC). 
It's worth noting that our framework and label flip do not need to know the training set, as they use the validation set as a surrogate dataset. 
The effectiveness of all the compared algorithms is subsequently evaluated on a test set, never seen during the optimization.
We run our experimental evaluation five times, and we report the mean accuracy and the corresponding standard deviation.

In our experimental settings we first focus on binary classification tasks, as done in \cite{biggio_poisoning_2013, demontis_why_nodate}. To this end, we selected pairs of classes from the datasets at hand. Secondly, we propose some results in a multi-labels scenario, which \cite{biggio_poisoning_2013, demontis_why_nodate} have not been designed for.

We use the implementation provided in \cite{melis2019secml} to generate poisoning points for \cite{biggio_poisoning_2013, demontis_why_nodate}. Code for all experiments
can be found at \href{https://github.com/Cinofix/beta_poisoning}{https://github.com/Cinofix/beta\_poisoning}.

\subsection{Digit Recognition} ~ \label{mnist_exp}
We consider the problem of digits recognition using the MNIST dataset, containing gray-scale $28\times 28$px images for $10$ classes (from 0 to 9). Each digit image so consists of $784$ pixels ranging from $0$ to $255$. We normalize pixels by dividing their values by $255$ and we use them as our features. We focus on the two-digits recognition problem, considering pairs $9$ vs. $8$ and $4$ vs. $0$ (as in \cite{biggio_poisoning_2013}). We randomly sample $400, 1000, 1000$ samples for training, validation, and test set for each pair of digits. This setting is similar to the one reported in \cite{biggio_poisoning_2013}, but we increased the number of training and validation samples.

We reported in Figure \ref{fig:svm_mnist} and Figure \ref{fig:logistic_mnist} the results obtained for SVM and LC with regularization parameter $C=1$ and $C=100$.
\begin{figure*}[t]
    \centering
    \includegraphics[width=0.45\textwidth]{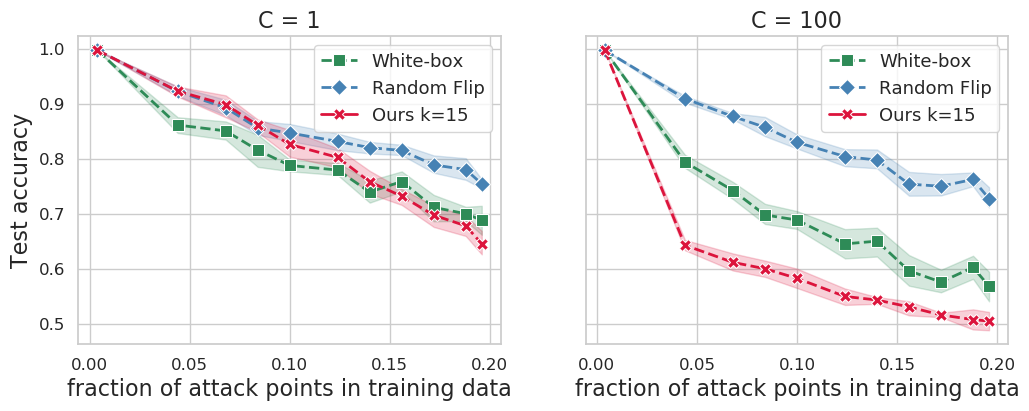}
    \includegraphics[width=0.45\textwidth]{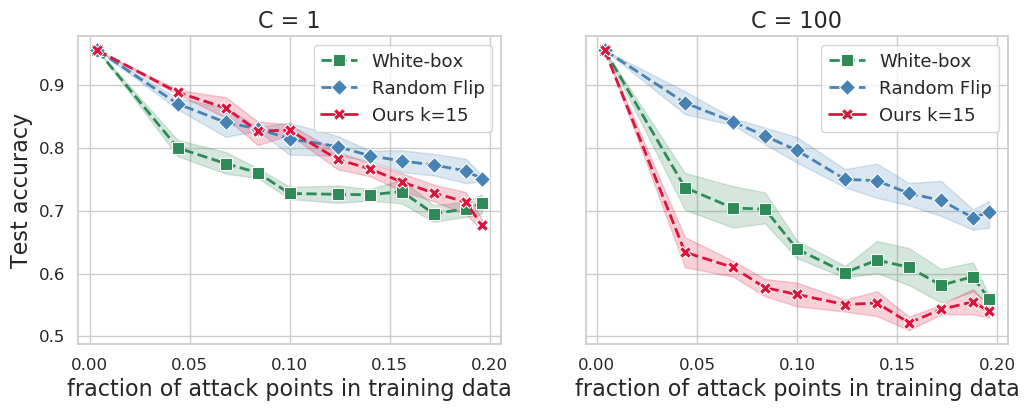}
  \caption{Accuracy on the test set for SVM, with regularization $C=1$ and $C=100$, under poisoning attack. (Left) results for MNIST pair 4 vs. 0, (right) results for pair 9 vs. 8. In green the performance for \cite{biggio_poisoning_2013}.}
  \label{fig:svm_mnist}
\end{figure*}

\begin{figure*}[t]
    \centering
    \includegraphics[width=0.45\textwidth]{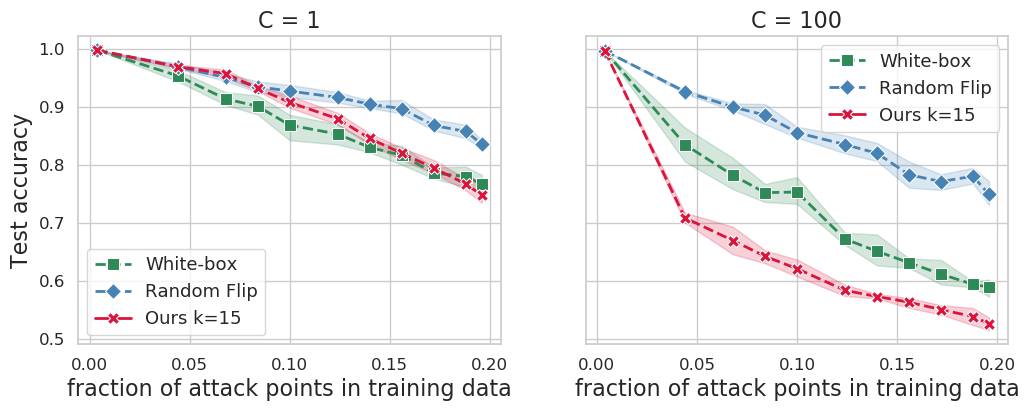}
    \includegraphics[width=0.45\textwidth]{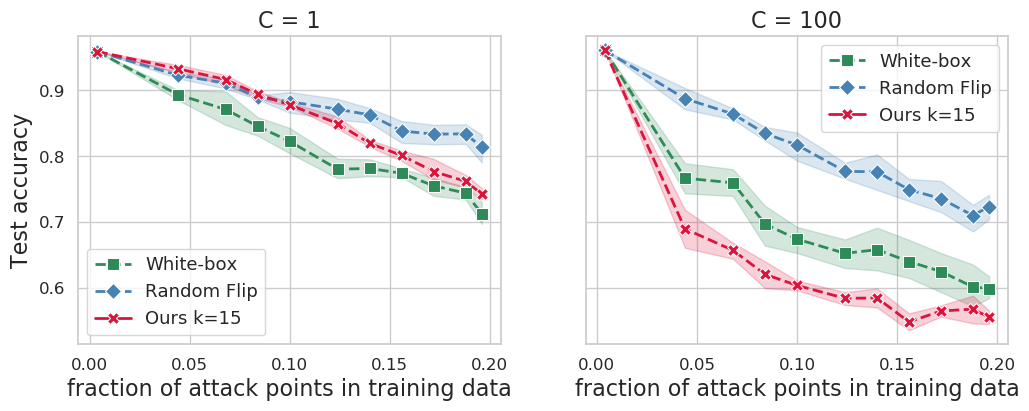}
  \caption{Accuracy on the test set for LC, with regularization $C=1$ and $C=100$, under poisoning attack. (Left) results for MNIST pair 4 vs. 0, (right) results for pair 9 vs. 8. In green the performance for \cite{demontis_why_nodate}.}
  \label{fig:logistic_mnist}
\end{figure*}

One can clearly see a steady growth of the attack effectiveness with the increasing fraction of poisoning points added to the training set. In particular, we observe that when the penalty term $C$, increases the target models become less robust against poisoning attacks. 
These results have also been observed in \cite{demontis_why_nodate}, where the authors state that strongly regularized classifiers tend to have smaller input gradients, i.e., they learn smoother functions that are more robust to attacks. Notably, the performance, in terms of attacker's objective, obtained by our framework are comparable with \cite{biggio_poisoning_2013}, or even better when the regularization of the target models decreases. Table \ref{table:mnist_time} reports the computational costs needed to run the three poisoning algorithms when the percentage of attack points in the training set is $20\%$.

\subsection{CIFAR-10 Images Recognition} ~ \label{cifar_exp}
This section presents the results obtained with the CIFAR-10 dataset, containing $32\times 32$px colors images. We normalize the pixels as in Section \ref{mnist_exp}, and we use them as features. Compared to previously described experiments on the digit recognition task, each sample has more features, $3072$ rather than $784$. 
We perform the experiments on the CIFAR-10 dataset to show how our poisoning algorithm scales on a larger dataset compared to \cite{biggio_poisoning_2013, demontis_why_nodate}.
We consider the two pairs of classes with the highest accuracy on untainted dataset, \textit{frog} vs. \textit{ship} and \textit{horse} vs. \textit{ship}. We randomly sample, for each of them, $300, 500$ and $1000$ images to build our training, validation and test set, respectively. 
We report in Figures \ref{fig:svm_cifar} and \ref{fig:logistic_cifar} the results of poisoning against linear SVM and LC with different regularization strengths. 
Our results on the CIFAR-10 dataset are consistent with those described in Section \ref{mnist_exp}. This means that our poisoning samples are effective even on large datasets. Notably, the computational gap for LC, reported in Table \ref{table:cifar_time}, is significantly increased, favoring our method.

\begin{figure*}[t]
    \centering
		\includegraphics[width=0.45\textwidth]{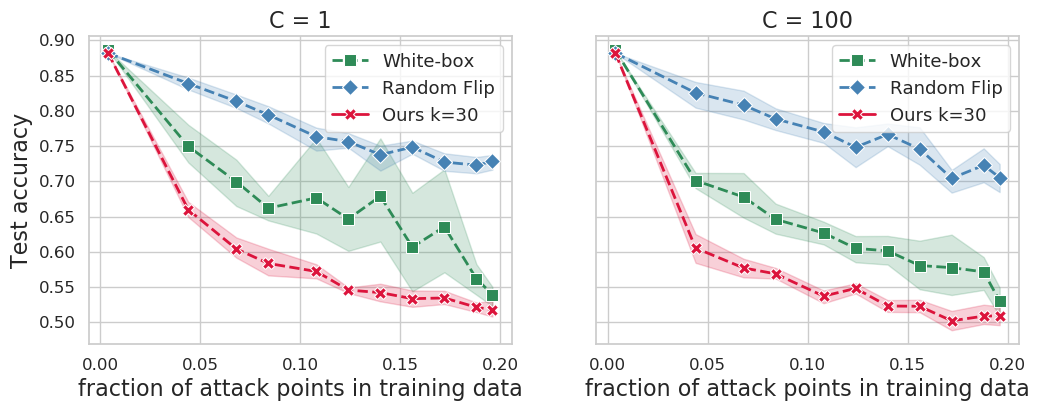}
        \includegraphics[width=0.45\textwidth]{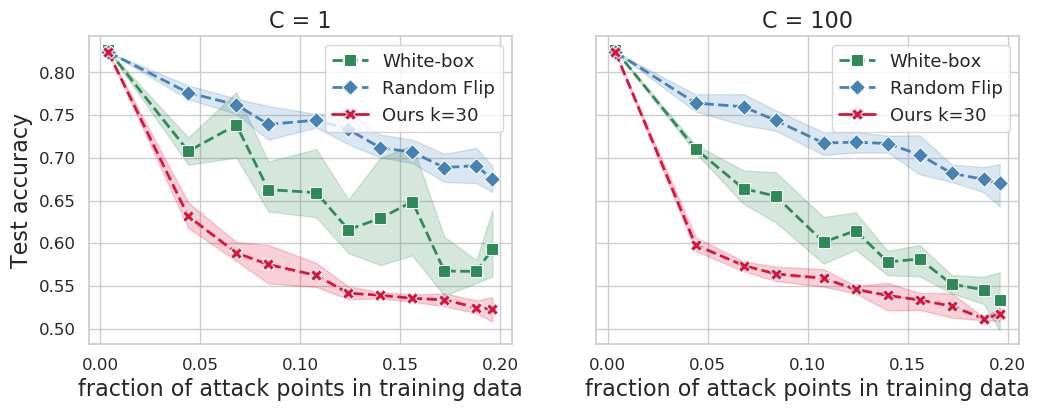}
  \caption{Accuracy on the test set for SVM, with regularization $C=1$ and $C=100$, under poisoning attack. (Left) results for CIFAR-10 pair frog vs. ship, (right) results for pair horse vs. ship. In green the performance for \cite{biggio_poisoning_2013}.}
  \label{fig:svm_cifar}
\end{figure*}

\begin{figure*}[t]
    \centering
    \includegraphics[width=0.45\textwidth]{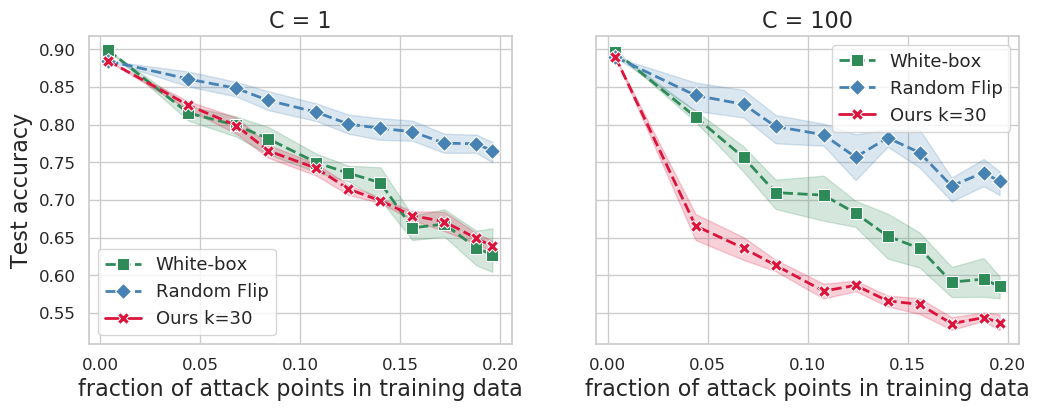}
    \includegraphics[width=0.45\textwidth]{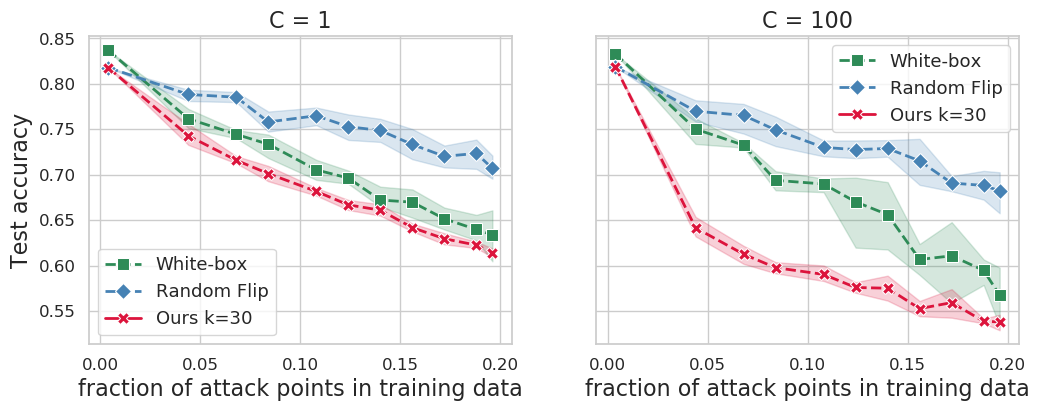}
  \caption{Accuracy on the test set for LC, with regularization $C=1$ and $C=100$, under poisoning attack. (Left) results for CIFAR-10 pair frog vs. ship, (right) results for pair horse vs. ship. In green the performance for \cite{demontis_why_nodate}.}
  \label{fig:logistic_cifar}
\end{figure*}


\subsection{Multiple labels} ~ \label{mnist_triplet}
In this section, we propose some results of our poisoning algorithm against non-binary classifiers. We consider two random triplet of classes from MNIST $(\{3,7,5\} \text{ and } \{9,4,0\})$. For each triplet we created training, validation and test set by randomly sampling $400$, $1000$, $1000$ images. We test the effectiveness of poisoning against an SVM with regularization term $C=1$ and $C=100$.
We observe that \cite{biggio_poisoning_2013} is only suited for binary classification problems, so for that reason, no comparison with that algorithm is provided.
\begin{figure*}[t]
    \centering
    \includegraphics[width=0.45\textwidth]{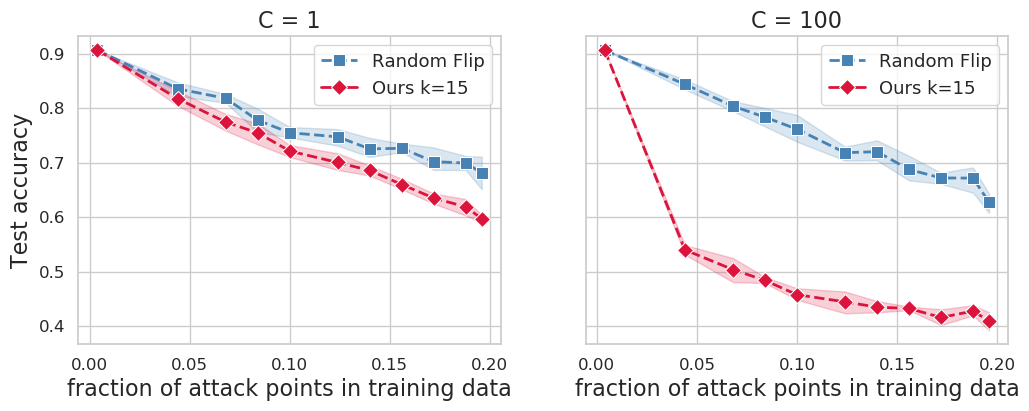}
    \includegraphics[width=0.45\textwidth]{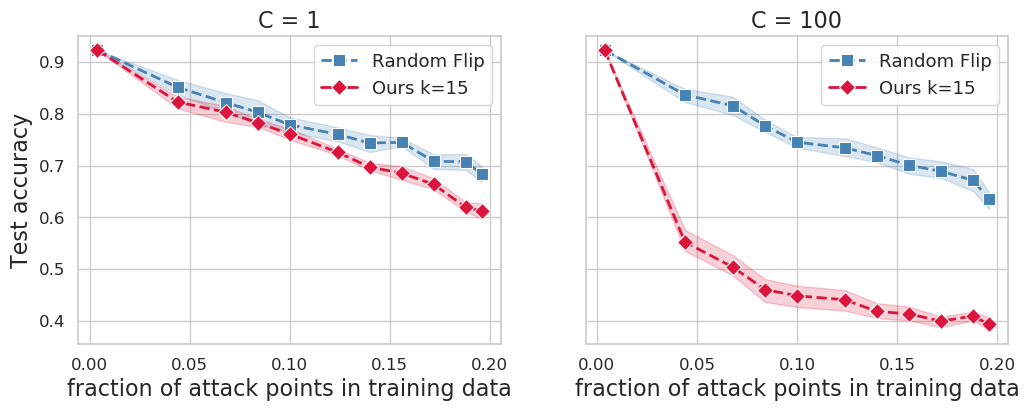}

  \caption{Accuracy on the test set for SVM, with regularization $C=1$ and $C=100$, under poisoning attack. The two on the left are obtained on triplet $\{3,7,5\}$, and the two on the right are obtained for triplet $\{9,4,0\}$.}
  \label{fig:svm_triplet}
\end{figure*}

Figure \ref{fig:svm_triplet} clearly emphasizes our poisoning algorithm's effectiveness against multi-labels linear classifier. We observe that when the regularization term $C$ increases, the robustness to poisoning decreased significantly. Notably, the two random triplets' performance has the same trend, confirming the goodness and stability of our approach to random selection.
The comparison with the label flip attack proves that the proposed algorithm \textit{counter-intuitive} is effective even against multi-labels classification tasks.

\subsection{Time Comparison}
This section analyzes the computational costs provided by our approach, and the white-box threat algorithms \cite{biggio_poisoning_2013, demontis_why_nodate}. We run our experiments on a Intel$^\text{\textregistered}$ Xeon$^\text{\textregistered}$ Processor E5-2690 v3. 

Table \ref{table:mnist_time} shows the results obtained for the two MNIST pairs when generating $100$ poisoning samples($20\%$ of the training set is poisoned). As we expected, the comparison of the attack times shows the proposed algorithm's reduced computational cost, highlighting a significant gap. Even if, as shown in Section \ref{mnist_exp}, the attack effectiveness is comparable, we can craft poisoning points more efficiently without solving a complex bi-level optimization problem. Moreover, our experimental analysis highlights the running time of the attack that solves the bi-level optimization problem is strongly influenced by different factors, such as the learning algorithm and the regularization strength. That attack is slower against LC than against SVM and is less computationally expensive against strongly regularized classifiers. Whereas the running time of the proposed algorithm is almost constant.

\begin{table}[t]
\centering
\caption{Computational cost comparison between poisoning algorithms against SVM and LC for MNIST 4-0 and 9-8.}
{\def\arraystretch{1.4}\tabcolsep=5px

\begin{tabular}{ccccc}
\hline
\multirow{2}{*}{Model}                  & \multirow{2}{*}{Dataset} & \multirow{2}{*}{Generator} & \multicolumn{2}{c}{Time in s}   \\ \cline{4-5} 
 & & & C=1 & \multicolumn{1}{l}{C=100} \\ \hline
\multirow{4}{*}{SVM} & 4-0 & \cite{biggio_poisoning_2013} &  $123.44 \pm 11.59$   & \multicolumn{1}{l}{}  $148.99 \pm 44.66 $\\
& 4-0 & \textbf{Ours} &  $\mathbf{11.06 \pm 0.60}$   & $\mathbf{11.40 \pm 1.78}$\\
& 9-8 & \cite{biggio_poisoning_2013} & $132.39 \pm 24.64$ & $168.17 \pm 53.26$ \\
& 9-8 & \textbf{Ours} &  $\mathbf{11.09 \pm 0.25}$   & $\mathbf{11.27 \pm 0.22}$\\ \hline

\multicolumn{1}{c}{\multirow{4}{*}{LC}} & 4-0 & \cite{demontis_why_nodate} &  $261.46 \pm 30.69$ & \multicolumn{1}{l}{}  $459.57 \pm 22.60$    \\
\multicolumn{1}{l}{} & 4-0 & \textbf{Ours} & $ \mathbf{10.70 \pm 0.55} $ & $11.06 \pm 1.64$ \\
\multicolumn{1}{l}{} & 9-8 & \cite{demontis_why_nodate} &  $285.09 \pm 31.39$   & $458.22 \pm 15.55$\\
\multicolumn{1}{l}{} & 9-8 & \textbf{Ours} & $\mathbf{11.17 \pm 0.21}$ & $\mathbf{11.45 \pm 0.48}$\\ \hline
\end{tabular}
}
\label{table:mnist_time}
\end{table}

Similarly, Table \ref{table:cifar_time} reports the computational costs for the two CIFAR-10 pairs when generating $75$ poisoning samples ($20\%$ of the training set is poisoned). Even in this case, our algorithm best perform with significant evidence against SVM and LC. We notice that the computational performance gap between SVM and LC is higher than the one exhibited on the MNIST dataset. 
On the CIFAR-10 dataset, when the classifiers complexity increases, the performance for \cite{biggio_poisoning_2013} and \cite{demontis_why_nodate} improves, but not enough to bridge the gap with our algorithm.

\begin{table}[t]
\centering
\caption{Computational cost comparison between poisoning algorithms against SVM and LC for CIFAR-10 frog-ship and horse-ship.}

{\def\arraystretch{1.4}\tabcolsep=4px
\begin{tabular}{ccccc}
\hline
\multirow{2}{*}{Model}                  & \multirow{2}{*}{Dataset} & \multirow{2}{*}{Generator} & \multicolumn{2}{c}{Time in s}   \\ \cline{4-5} 
 & & & C=1 & \multicolumn{1}{l}{C=100} \\ \hline
\multirow{4}{*}{SVM} & frog-ship & \cite{biggio_poisoning_2013} &  $115.72 \pm 24.72$   &  \multicolumn{1}{l}{}\noindent $40.21 \pm 2.35 $\\ 
& frog-ship & \textbf{Ours} &  $\mathbf{06.95 \pm 0.31}$   & $\mathbf{6.77 \pm 0.54}$\\

& horse-ship & \cite{biggio_poisoning_2013} & $79.05 \pm 31.04$ & $46.90 \pm 05.68$ \\
& horse-ship & \textbf{Ours} &  $\mathbf{06.41 \pm 0.52}$   & $\mathbf{6.46  \pm 0.38}$\\ \hline

\multicolumn{1}{c}{\multirow{4}{*}{LC}} & frog-ship & \cite{demontis_why_nodate} &  $16851.3 \pm 2506.5$ & \multicolumn{1}{l}{}  $6386.32 \pm 1076.1$\\
\multicolumn{1}{l}{} & frog-ship & \textbf{Ours} & $ \mathbf{08.89 \pm 0.48} $ & $\mathbf{08.82 \pm 0.68}$ \\

\multicolumn{1}{l}{} & horse-ship & \cite{demontis_why_nodate} &  $17788.9 \pm 1335.1$   & $7029.8 \pm 343.08$\\
\multicolumn{1}{l}{} & horse-ship & \textbf{Ours} & $\mathbf{06.73 \pm 0.71}$ & $\mathbf{06.58 \pm 0.44}$\\ \hline
\end{tabular}
}
\label{table:cifar_time}
\end{table}

\subsection{Ablation Study}
In this section, we study our poisoning algorithm's effectiveness by varying the number of prototypes, the cardinality of $\set S$. We remark that the number of prototypes corresponds exactly with the number of coefficients $\boldsymbol \beta$ to be optimized during the learning process. We use the same configuration of MNIST (4 vs. 0) and CIFAR-10 (frog vs. ship), detailed in Section \ref{mnist_exp} and \ref{cifar_exp}. We let the number of prototypes vary from $2$ to $30$ and analyze our poisoning attack's performance against a linear SVM with regularization term $C=1$.

\begin{figure}[]
    \centering
    \includegraphics[width=0.45\textwidth]{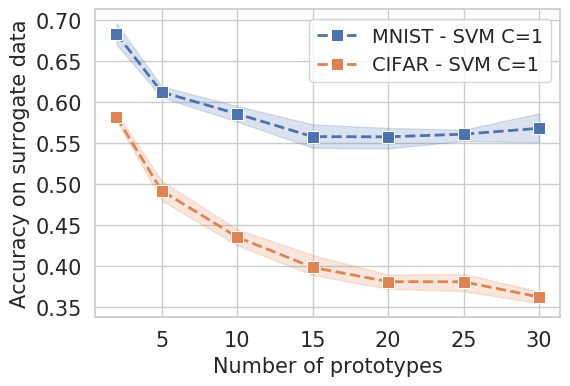}
  \caption{Effectiveness of poisoning when ~15\% of the dataset is poisoned. The x-axis represents the number of prototypes in $\set S$, equal to the number of optimized variables. The y-axis shows the accuracy of the system under attack on a validation/surrogate dataset.}
  \label{fig:ablation}
\end{figure}

Figure \ref{fig:ablation} shows that the number of prototypes chosen may significantly influence our algorithm's performance. In particular, when increasing it, our framework seems to create more powerful poisoning points. We observe for a smaller dataset like MNIST, 15 prototypes are sufficient. Conversely, for a more complex dataset like CIFAR-10, the optimal number of prototypes is 30. This reduction represents a significant improvement; indeed if we consider Problem (\ref{eq:poisoning_problem_outer})-(\ref{eq:poisoning_problem_box}) and Problem \ref{beta_poisoning}, the number of variables to optimize is equal to the sample's dimension. Thanks to our variable reduction trick, we can optimize only 15 out of 784 variables for MNIST and 30 out of 3072 for CIFAR-10. These results indicate that this approach is probably practicable also on datasets with many more features than the CIFAR-10 dataset.

\section{Related Work} ~ \label{sec:related_work}
Evasion and data poisoning attacks represent critical challenges to the design of novel AI systems. \\

\noindent \textbf{Evasion attacks.} Evasion attacks aim to find adversarial inputs misclassified at test time \cite{carlini_towards_2017, frederickson_attack_2018}. 
However, in the presence of supervised conditions, such as staffed security, an adversary may not be able to alter the input sample as required for evasion threats \cite{shafahi_poison_2018}. In such scenarios, the attacker should trick the models before it is deployed. In this regard, integrity data poisoning attacks lead to a change in the target model's internal parameters to meet a specific attacker's objective at test time.  

\noindent \textbf{Integrity data poisoning.} Integrity data poisoning attacks allow the attacker to leverage backdoor, injected in the training set, for his malicious purpose \cite{geiping_witches_2020, zhu_transferable_2019, shafahi_poison_2018, chen_targeted_2017}.
The essential property of integrity violation is that the model's overall performance is not affected, but only a few target samples are misclassified.

\noindent\textbf{Availability data poisoning.} Availability data poisoning attacks aim to induce the system to a Denial-of-Service (DoS), limiting its use to all authorized users. In \cite{biggio_poisoning_2013} the authors proposed the first availability poisoning algorithm. That algorithm was tailored against support vector machines to create effective poisoning samples in a white-box scenario. Later on, in \cite{demontis_why_nodate} the same approach was adapted and exploit to trick a logistic regression classifier.
In \cite{munoz-gonzalez_towards_2017} the authors remarked that solving the bi-level formulation of Problem, (\ref{eq:poisoning_problem_outer})-(\ref{eq:poisoning_problem_box}) requires the inversion of the Hessian matrix of the model's parameters. Therefore, the algorithm to solve this problem exactly has a cubic complexity with respect to the number of parameters. This aspect makes the usage of this algorithm computationally prohibitive for a variety of practical settings. Therefore, the authors proposed an algorithm that finds an approximated solution to that problem using a technique called back-gradient optimization. They showed that this algorithm allows attacking small deep neural networks efficiently. 
The authors of \cite{demontis_why_nodate} observed that the poisoning samples, crafted to poison a target model, are effective even against other systems. This transferability property allows the adversary to generate attacks even if she does not directly know the target model.
\cite{yang_generative_2017} proposed a gradient method to estimate the gradient of the poisoning sample efficiently. However, the authors observed that this solution does not scale with the number of features. Therefore, to mitigate this issue,  they propose a generative approach, which still requires re-training the target model at each iteration.

As reported in Section \ref{par:poison_strategy} both availability and integrity data poisoning can be expressed with a bi-level optimization formulation. Nevertheless, for integrity violations, more prominent and straightforward heuristics have been proposed to solve this problem. For instance, in \cite{shafahi_poison_2018} and \cite{zhu_transferable_2019} the authors craft the poison samples so that they collide or surround the targeted image in feature space. They observed that their re-formulations allows them to obtain effective and computationally efficient results. 
The literature for availability data poisoning lacks similar and efficient strategies, limiting its application for large-scale problems.

\noindent \textbf{Clustering availability violation.}
Causative attacks have attracted attention even in unsupervised settings. Indeed, authors in \cite{battista_is_data_clustering, black_box_clustering, poison_malware_clustering} test the robustness of clustering-based applications against availability data poisoning attacks. They observed that even unsupervised algorithms, such as clustering, are not safe against malicious users.

\section{Conclusion}

In this paper, we addressed the question: do we really need to use an algorithm that solves the bi-level problem exactly (the ``hammer'') to carry out an availability poisoning attack against a simple linear classifier (the ``nut'')?  Our analysis shows that, for this class of problems, we don't really need it. Indeed, we evince that our heuristic attack reaches comparable, or even better, results compared to theoretical and time-consuming formulations. We propose a re-parametrization trick to reduce the number of variables during the learning process. We compared the computational costs of the proposed algorithm with the ones obtained for ``hammer-based'' poisoning algorithms, namely the ones that solve exactly the bi-level optimization problem. We provided experimental evidences that we can poison target models with a significantly lower computational cost. Our approach may open the door toward the design of more efficient heuristics to deceive and test critical systems against availability data poisoning. For future works, we aim to investigate approaches to extend our approach's against non-linear classifiers.

\section{Acknowledgment}
This work has been partially supported by the PRIN 2017 project RexLearn (grant no. 2017TWNMH2), funded by the Italian Ministry of Education, University and Research; and by BMK, BMDW, and the Province of Upper Austria in the frame of the COMET Programme managed by FFG in the COMET Module S3AI.

\bibliographystyle{IEEEtran}
\bibliography{IEEEabrv, ijcai21}

\begin{thebibliography}{10}
\providecommand{\url}[1]{#1}
\csname url@samestyle\endcsname
\providecommand{\newblock}{\relax}
\providecommand{\bibinfo}[2]{#2}
\providecommand{\BIBentrySTDinterwordspacing}{\spaceskip=0pt\relax}
\providecommand{\BIBentryALTinterwordstretchfactor}{4}
\providecommand{\BIBentryALTinterwordspacing}{\spaceskip=\fontdimen2\font plus
\BIBentryALTinterwordstretchfactor\fontdimen3\font minus
  \fontdimen4\font\relax}
\providecommand{\BIBforeignlanguage}[2]{{%
\expandafter\ifx\csname l@#1\endcsname\relax
\typeout{** WARNING: IEEEtran.bst: No hyphenation pattern has been}%
\typeout{** loaded for the language `#1'. Using the pattern for}%
\typeout{** the default language instead.}%
\else
\language=\csname l@#1\endcsname
\fi
#2}}
\providecommand{\BIBdecl}{\relax}
\BIBdecl

\bibitem{barreno_security_2010}
M.~Barreno, B.~Nelson, A.~D. Joseph, and J.~D. Tygar, ``The security of machine
  learning,'' \emph{Mach. Learn.}, vol.~81, no.~2, pp. 121--148, 2010.

\bibitem{biggio_evasion_2013}
B.~Biggio, I.~Corona, D.~Maiorca, B.~Nelson, N.~Srndic, P.~Laskov, G.~Giacinto,
  and F.~Roli, ``Evasion attacks against machine learning at test time,'' in
  \emph{Machine Learning and Knowledge Discovery in Databases - European
  Conference, {ECML} {PKDD} 2013, Prague, Czech Republic, September 23-27,
  2013, Proceedings, Part {III}}, ser. Lecture Notes in Computer Science,
  H.~Blockeel, K.~Kersting, S.~Nijssen, and F.~Zelezn{\'{y}}, Eds., vol.
  8190.\hskip 1em plus 0.5em minus 0.4em\relax Springer, 2013, pp. 387--402.

\bibitem{su_one_2019}
J.~Su, D.~V. Vargas, and K.~Sakurai, ``One pixel attack for fooling deep neural
  networks,'' \emph{{IEEE} Trans. Evol. Comput.}, vol.~23, no.~5, pp. 828--841,
  2019.

\bibitem{carlini_towards_2017}
N.~Carlini and D.~A. Wagner, ``Towards evaluating the robustness of neural
  networks,'' in \emph{2017 {IEEE} Symposium on Security and Privacy, {SP}
  2017, San Jose, CA, USA, May 22-26, 2017}.\hskip 1em plus 0.5em minus
  0.4em\relax {IEEE} Computer Society, 2017, pp. 39--57.

\bibitem{goodfellow_explaining_2015}
I.~J. Goodfellow, J.~Shlens, and C.~Szegedy, ``Explaining and harnessing
  adversarial examples,'' in \emph{3rd International Conference on Learning
  Representations, {ICLR} 2015, San Diego, CA, USA, May 7-9, 2015, Conference
  Track Proceedings}, Y.~Bengio and Y.~LeCun, Eds., 2015.

\bibitem{szegedy_properties}
C.~Szegedy, W.~Zaremba, I.~Sutskever, J.~Bruna, D.~Erhan, I.~J. Goodfellow, and
  R.~Fergus, ``Intriguing properties of neural networks,'' in \emph{2nd
  International Conference on Learning Representations, {ICLR} 2014, Banff, AB,
  Canada, April 14-16, 2014, Conference Track Proceedings}, Y.~Bengio and
  Y.~LeCun, Eds., 2014.

\bibitem{song18-cvpr}
K.~Eykholt, I.~Evtimov, E.~Fernandes, B.~Li, A.~Rahmati, C.~Xiao, A.~Prakash,
  T.~Kohno, and D.~Song, ``Robust physical-world attacks on deep learning
  visual classification,'' in \emph{2018 {IEEE} Conference on Computer Vision
  and Pattern Recognition, {CVPR} 2018, Salt Lake City, UT, USA, June 18-22,
  2018}.\hskip 1em plus 0.5em minus 0.4em\relax {IEEE} Computer Society, 2018,
  pp. 1625--1634.

\bibitem{demetrio20-arxiv-blackbox}
\BIBentryALTinterwordspacing
L.~Demetrio, B.~Biggio, G.~Lagorio, F.~Roli, and A.~Armando,
  ``Functionality-preserving {Black}-box {Optimization} of {Adversarial}
  {Windows} {Malware},'' \emph{arXiv:2003.13526 [cs]}, Sep. 2020, arXiv:
  2003.13526. [Online]. Available: \url{http://arxiv.org/abs/2003.13526}
\BIBentrySTDinterwordspacing

\bibitem{shafahi_poison_2018}
A.~Shafahi, W.~R. Huang, M.~Najibi, O.~Suciu, C.~Studer, T.~Dumitras, and
  T.~Goldstein, ``Poison frogs! targeted clean-label poisoning attacks on
  neural networks,'' in \emph{Advances in Neural Information Processing Systems
  31: Annual Conference on Neural Information Processing Systems 2018, NeurIPS
  2018, December 3-8, 2018, Montr{\'{e}}al, Canada}, S.~Bengio, H.~M. Wallach,
  H.~Larochelle, K.~Grauman, N.~Cesa{-}Bianchi, and R.~Garnett, Eds., 2018, pp.
  6106--6116.

\bibitem{industry_perspective}
R.~S.~S. Kumar, M.~Nystr{\"{o}}m, J.~Lambert, A.~Marshall, M.~Goertzel,
  A.~Comissoneru, M.~Swann, and S.~Xia, ``Adversarial machine learning-industry
  perspectives,'' in \emph{2020 {IEEE} Security and Privacy Workshops, {SP}
  Workshops, San Francisco, CA, USA, May 21, 2020}.\hskip 1em plus 0.5em minus
  0.4em\relax {IEEE}, 2020, pp. 69--75.

\bibitem{LearningfromTaysBlog}
``Learning from tay’s introduction - the official microsoft blog,''
  https://blogs.microsoft.com/blog/2016/03/25/learning-tays-introduction/,
  accessed: 2020-11-21 03:54:49.

\bibitem{chen_targeted_2017}
X.~Chen, C.~Liu, B.~Li, K.~Lu, and D.~Song, ``Targeted backdoor attacks on deep
  learning systems using data poisoning,'' \emph{CoRR}, vol. abs/1712.05526,
  2017.

\bibitem{geiping_witches_2020}
J.~Geiping, L.~Fowl, W.~R. Huang, W.~Czaja, G.~Taylor, M.~Moeller, and
  T.~Goldstein, ``Witches' brew: Industrial scale data poisoning via gradient
  matching,'' \emph{CoRR}, vol. abs/2009.02276, 2020.

\bibitem{munoz-gonzalez_towards_2017}
L.~Mu{\~{n}}oz{-}Gonz{\'{a}}lez, B.~Biggio, A.~Demontis, A.~Paudice,
  V.~Wongrassamee, E.~C. Lupu, and F.~Roli, ``Towards poisoning of deep
  learning algorithms with back-gradient optimization,'' in \emph{Proceedings
  of the 10th {ACM} Workshop on Artificial Intelligence and Security, AISec@CCS
  2017, Dallas, TX, USA, November 3, 2017}, B.~M. Thuraisingham, B.~Biggio,
  D.~M. Freeman, B.~Miller, and A.~Sinha, Eds.\hskip 1em plus 0.5em minus
  0.4em\relax {ACM}, 2017, pp. 27--38.

\bibitem{demontis_why_nodate}
A.~Demontis, M.~Melis, M.~Pintor, M.~Jagielski, B.~Biggio, A.~Oprea,
  C.~Nita{-}Rotaru, and F.~Roli, ``Why do adversarial attacks transfer?
  explaining transferability of evasion and poisoning attacks,'' in \emph{28th
  {USENIX} Security Symposium, {USENIX} Security 2019, Santa Clara, CA, USA,
  August 14-16, 2019}, N.~Heninger and P.~Traynor, Eds.\hskip 1em plus 0.5em
  minus 0.4em\relax {USENIX} Association, 2019, pp. 321--338.

\bibitem{biggio_poisoning_2013}
B.~Biggio, B.~Nelson, and P.~Laskov, ``Poisoning attacks against support vector
  machines,'' in \emph{Proceedings of the 29th International Conference on
  Machine Learning, {ICML} 2012, Edinburgh, Scotland, UK, June 26 - July 1,
  2012}.\hskip 1em plus 0.5em minus 0.4em\relax icml.cc / Omnipress, 2012.

\bibitem{koh_understanding_nodate}
P.~W. Koh and P.~Liang, ``Understanding black-box predictions via influence
  functions,'' in \emph{Proceedings of the 34th International Conference on
  Machine Learning, {ICML} 2017, Sydney, NSW, Australia, 6-11 August 2017},
  ser. Proceedings of Machine Learning Research, D.~Precup and Y.~W. Teh, Eds.,
  vol.~70.\hskip 1em plus 0.5em minus 0.4em\relax {PMLR}, 2017, pp. 1885--1894.

\bibitem{yang_generative_2017}
C.~Yang, Q.~Wu, H.~Li, and Y.~Chen, ``Generative poisoning attack method
  against neural networks,'' \emph{CoRR}, vol. abs/1703.01340, 2017.

\bibitem{biggio_wild_2018}
B.~Biggio and F.~Roli, ``Wild patterns: Ten years after the rise of adversarial
  machine learning,'' \emph{Pattern Recognit.}, vol.~84, pp. 317--331, 2018.

\bibitem{transfer_iclr17}
Y.~Liu, X.~Chen, C.~Liu, and D.~Song, ``Delving into transferable adversarial
  examples and black-box attacks,'' in \emph{5th International Conference on
  Learning Representations, {ICLR} 2017, Toulon, France, April 24-26, 2017,
  Conference Track Proceedings}.\hskip 1em plus 0.5em minus 0.4em\relax
  OpenReview.net, 2017.

\bibitem{transfer_cvpr19}
C.~Xie, Z.~Zhang, Y.~Zhou, S.~Bai, J.~Wang, Z.~Ren, and A.~L. Yuille,
  ``Improving transferability of adversarial examples with input diversity,''
  in \emph{{IEEE} Conference on Computer Vision and Pattern Recognition, {CVPR}
  2019, Long Beach, CA, USA, June 16-20, 2019}.\hskip 1em plus 0.5em minus
  0.4em\relax Computer Vision Foundation / {IEEE}, 2019, pp. 2730--2739.

\bibitem{DBLP:journals/datamine/Burges98}
C.~J.~C. Burges, ``A tutorial on support vector machines for pattern
  recognition,'' \emph{Data Min. Knowl. Discov.}, vol.~2, no.~2, pp. 121--167,
  1998.

\bibitem{taheri_defending_2020}
R.~Taheri, R.~Javidan, M.~Shojafar, Z.~Pooranian, A.~Miri, and M.~Conti, ``On
  defending against label flipping attacks on malware detection systems,''
  \emph{Neural Comput. Appl.}, vol.~32, no.~18, pp. 14\,781--14\,800, 2020.

\bibitem{battista_is_data_clustering}
B.~Biggio, I.~Pillai, S.~R. Bul{\`{o}}, D.~Ariu, M.~Pelillo, and F.~Roli, ``Is
  data clustering in adversarial settings secure?'' in \emph{AISec'13,
  Proceedings of the 2013 {ACM} Workshop on Artificial Intelligence and
  Security, Co-located with {CCS} 2013, Berlin, Germany, November 4, 2013},
  A.~Sadeghi, B.~Nelson, C.~Dimitrakakis, and E.~Shi, Eds.\hskip 1em plus 0.5em
  minus 0.4em\relax {ACM}, 2013, pp. 87--98.

\bibitem{bengio_label_flip}
\BIBentryALTinterwordspacing
C.~Zhang, S.~Bengio, M.~Hardt, B.~Recht, and O.~Vinyals, ``Understanding deep
  learning requires rethinking generalization,'' in \emph{5th International
  Conference on Learning Representations, {ICLR} 2017, Toulon, France, April
  24-26, 2017, Conference Track Proceedings}.\hskip 1em plus 0.5em minus
  0.4em\relax OpenReview.net, 2017. [Online]. Available:
  \url{https://openreview.net/forum?id=Sy8gdB9xx}
\BIBentrySTDinterwordspacing

\bibitem{melis2019secml}
M.~Melis, A.~Demontis, M.~Pintor, A.~Sotgiu, and B.~Biggio, ``secml: A python
  library for secure and explainable machine learning,'' \emph{arXiv preprint
  arXiv:1912.10013}, 2019.

\bibitem{frederickson_attack_2018}
C.~Frederickson, M.~Moore, G.~Dawson, and R.~Polikar, ``Attack strength vs.
  detectability dilemma in adversarial machine learning,'' in \emph{2018
  International Joint Conference on Neural Networks, {IJCNN} 2018, Rio de
  Janeiro, Brazil, July 8-13, 2018}.\hskip 1em plus 0.5em minus 0.4em\relax
  {IEEE}, 2018, pp. 1--8.

\bibitem{zhu_transferable_2019}
C.~Zhu, W.~R. Huang, H.~Li, G.~Taylor, C.~Studer, and T.~Goldstein,
  ``Transferable clean-label poisoning attacks on deep neural nets,'' in
  \emph{Proceedings of the 36th International Conference on Machine Learning,
  {ICML} 2019, 9-15 June 2019, Long Beach, California, {USA}}, ser. Proceedings
  of Machine Learning Research, K.~Chaudhuri and R.~Salakhutdinov, Eds.,
  vol.~97.\hskip 1em plus 0.5em minus 0.4em\relax {PMLR}, 2019, pp. 7614--7623.

\bibitem{black_box_clustering}
\BIBentryALTinterwordspacing
A.~E. Cin{\`{a}}, A.~Torcinovich, and M.~Pelillo, ``A black-box adversarial
  attack for poisoning clustering,'' \emph{CoRR}, vol. abs/2009.05474, 2020.
  [Online]. Available: \url{https://arxiv.org/abs/2009.05474}
\BIBentrySTDinterwordspacing

\bibitem{poison_malware_clustering}
B.~Biggio, K.~Rieck, D.~Ariu, C.~Wressnegger, I.~Corona, G.~Giacinto, and
  F.~Roli, ``Poisoning behavioral malware clustering,'' in \emph{Proceedings of
  the 2014 Workshop on Artificial Intelligent and Security Workshop, AISec
  2014, Scottsdale, AZ, USA, November 7, 2014}, C.~Dimitrakakis, A.~Mitrokotsa,
  B.~I.~P. Rubinstein, and G.~Ahn, Eds.\hskip 1em plus 0.5em minus 0.4em\relax
  {ACM}, 2014, pp. 27--36.

\end{thebibliography}

\end{document}